\documentclass[letterpaper, 10 pt, conference]{ieeeconf}  

\IEEEoverridecommandlockouts                              

\usepackage[T1]{fontenc} 
\usepackage[utf8]{inputenc} 
\usepackage{graphicx}
\usepackage{color}
\usepackage{amssymb}
\usepackage{amsmath}

\title{\LARGE \bf
Utilizing Mood-Inducing Background Music in Human-Robot Interaction
}

\author{Elad Liebman$^{1}$ and Peter Stone$^{2}$
\thanks{$^{1}$Dr. Elad Liebman is with SparkCognition Research,
        Austin, TX 
        {\tt\small eliebman@sparkcognition.com}}%
\thanks{$^{2}$Prof.  Peter Stone is with the Computer Science Department, The University of Texas at Austin, and Sony AI,
        {\tt\small pstone@cs.utexas.edu}}%
}

\begin{document}

\maketitle
\thispagestyle{empty}
\pagestyle{empty}



%
\begin{abstract}
Past research has clearly established that music can affect mood and that mood affects emotional and cognitive processing, and thus decision-making. It follows that if a robot interacting with a person needs to predict the person's behavior, knowledge of the music the person is listening to when acting is a potentially relevant feature. To date, however, there has not been any concrete evidence that a robot can improve its human-interactive decision-making by taking into account what the person is listening to.  This research fills this gap by reporting the results of an experiment in which human participants were required to complete a task in the presence of an autonomous agent while listening to background music. Specifically, the participants drove a simulated car through an intersection while listening to music. The intersection was not empty, as another simulated vehicle, controlled autonomously, was also crossing the intersection in a different direction. Our results clearly indicate that such background information can be effectively incorporated in an agent's world representation in order to better predict people's behavior. We subsequently analyze how knowledge of music impacted both participant behavior and the resulting learned policy.\setcounter{footnote}{2}\footnote{An earlier version of part of the material in this paper appeared originally in the first author's Ph.D. Dissertation~\cite{liebman2020sequential} but it has not appeared in any pear-reviewed conference or journal.}

\end{abstract}

\section{Introduction}

Multiple studies in the field of cognitive psychology have established that mood-induced bias can affect how one processes information in a wide array of contexts and tasks \cite{elliott2002neural,olafson2001effects}. However, one important question has not been addressed in the literature -- can knowledge of the background stimulus, and specifically background music lead to better modeling of human behavior by a robot? Could the robot better predict people's behavior in this experiment given knowledge of what music the person is listening to? In this paper we set out to study this question.

In recent work, Liebman et al.\  illustrated how the emotional content of music informs the a priori expectation for the emotional content of verbal stimuli \cite{decbias}. Follow-up studies presented evidence for the complex impact of music-induced mood on risk-aware decision-making in the context of gambling behavior \cite{liebman2016impact},  and showed that music had a differential impact on decision behavior in the context of more complex cooperative tasks \cite{ISMIR2018}. However, the autonomous agent in the experiment described by Liebman et al. \cite{ISMIR2018} did not exhibit much agency in general. It always gave right of way to the human driver if the person reached the intersection first, and decided on its speed and intersection wait times uniformly at random in a rather conservative range. This fact raises the following question: what if the agent's behavior were replaced with a learning agent that was motivated by a goal similar to that of the subject's - to reach the other side of the intersection as quickly as possible without crashing? More importantly, could knowledge of the background music lead to better policy learning in this setting? Our main contribution is answering this question.

In this paper, we build on the work of Liebman et al. and design an experiment in which a person must control a car as it crosses a simulated intersection. This intersection is simultaneously being crossed by another vehicle, controlled by an autonomous agent, and therefore the person driving the simulated vehicle must reason about the autonomous vehicle's intentions (and vice versa). We study whether knowledge of the music background condition can be used by an autonomous agent to navigate the intersection more effectively. Our results suggest that taking the background music into account in the \emph{autonomous robot}'s world state representation can indeed lead to better policy learning.

\section{Background and Related Work}

Studies that induce mood, either through listening to happy/sad music or having participants write passages or see pictures based on a particular emotion, have shown mood-congruent impact on behavior across a range of tasks\cite{jeong2011congruence,de2002effectiveness,kuhbandner2013joint}. There is neurophysical evidence of music being strongly linked to brain regions linked with emotion and reward \cite{blood2001intensely}, and different musical patterns have been shown to have meaningful associations to emotional affectations \cite{paquette2013musical}. Not only is emotion a core part of music cognitive processing, it can also have a resounding impact on people's mental state, and aid in recovery, as shown for instance by Zumbansen et al. \cite{zumbansen2014combination}. People regularly use music to alter their moods, and evidence has been presented that music can alter the strength of emotional negativity bias \cite{chen2008music}.Some previous work has studied the general connection between gambling behavior and ambiance factors including music \cite{spenwyn2010role}. Additionally, Noseworthy and Finlay have studied the effects of music-induced dissociation and time perception in gambling establishments \cite{noseworthy2009comparison}. In the context of human-robot interaction, multiple studies have explicitly modeled emotional states of humans as part of their planning and learning architectures\cite{thomaz2016computational}. 

Little research has been done to quantitatively explore how music impacts the cooperative and adversarial behaviors of participants in social settings. Nayak et al. studied how music therapy impacted mood and social interaction in people recovering from brain injury \cite{nayak2000effect}. Greitemeyer presented evidence that exposure to music with prosocial lyrics reduces aggression \cite{greitemeyer2011exposure}. From a different perspective entirely, Baron was able to show how environmentally-induced mood helped improve negotiation and decrease adversarial behavior\cite{baron1990environmentally}. To the best of our knowledge, this paper is the first work to study how an \emph{autonomous robot} could leverage knowledge of background music and its impact on human behavior to learn a better interaction policy.

\section{Experimental Setup}
\label{exp_setup}

The main hypothesis of this research is that an autonomous robot can improve its human-interactive decision making by explicitly modeling the music being listened to by the person with whom it is interacting.  Such an effect might be observed in any setting such that a) the person's decisions are affected by background music; and 2) the agent's policy performance is affected by the person's decisions. Potential such settings range from stores, parks, and stadiums to elevators, parking garages, and online shopping and gaming. More specifically, it is highly relevant to autonomous vehicles driving in a mixed autonomy (some human drivers) setting.  In this paper, we focus on the latter task because it has been previously studied in such a setting, and because of its rising prominence outside academia. 

Specifically, this paper's experimental design is based on that used by Liebman et al.\ \cite{ISMIR2018}. Participants were given control of a simulated vehicle crossing an intersection. They had three control options - speed forward, go in reverse, and brake. In addition to the human-controlled vehicle, another vehicle, controlled autonomously by an artificial agent, was also crossing the intersection from a different direction. If the two cars collided, they would crash. Participants were instructed to safely cross the intersection without crashing. Participants were also instructed to do their best to reach the other side safely, and were warned the autonomous agent's behavior could be unpredictable. Throughout this paper, by ``waiting at the intersection'' we mean that either the person or the autonomous agent reached the intersection, then stopped for some nonzero wait time before proceeding. To offset any learning effects, each participant was given a about 3 minutes to interact with the system before  recording the session. 

\begin{figure}[!htb]
\begin{center}
 \includegraphics[width=\linewidth,height=120pt]{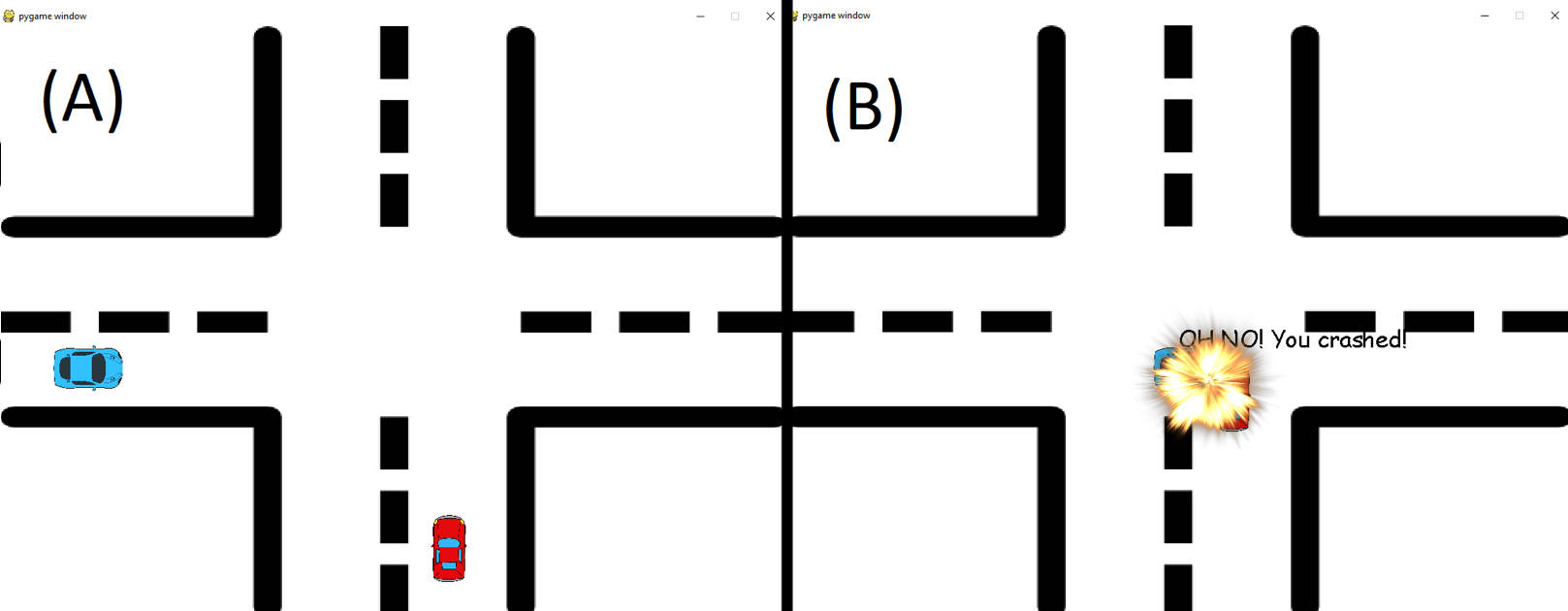}
 \caption{(A) A screen capture of the experiment. The red car was controlled by the participant. The blue car was controlled autonomously. A trial successfully ended when both vehicles reached the end of the screen. (B) A collision would result in a crash. After the crash, the trial terminated and the next trial began.}
 \label{chap5:fig1}
 \end{center}
\end{figure}

Each time both vehicles cleared the intersection and reached the end of the screen safely, the trial ended and a new trial commenced (a 3 second pause was introduced between trials). At the end of each trial, the task completion times (i.e. time to reach the end of the intersection) of both the human participant and the autonomous agent were shown. The experiment was divided into 16 blocks of 12 trials (for a total of 192 trials). In each block, a different song was played, alternating between positive and negative music across blocks. The order of the songs was counterbalanced across subjects and blocks. A 3 second pause before the beginning of each block was introduced to make sure the new song had started before a new trial commenced. Each experiment lasted approximately 30 minutes.  A snapshot of the experiment is presented in Figure \ref{chap5:fig1}.

%

\subsection{Participants \& Background Music}

For this experiment we collected data from 22 participants. All participants were graduate students who volunteered to participate in the study. In this experiment the focus was on the behavior of the autonomous agent, rather than the human participants, a fact which enabled us to aggregate data across participants, allowing for larger samples per condition compared to the previous experiment. 

To make sure the results presented by Liebman et al. were replicable, the music used for this experiment is the same as that which was used by Liebman et al. In their studies, Liebman et al. surveyed and isolated two clear types of music - music that is characterized by slow tempo, minor keys and somber tones, typical to traditionally ``sad'' music, and music that has upbeat tempo, major scales and colorful tones, which are traditionally considered to be typical of ``happy'' music. They proceeded to show that this music indeed inspired the expected mood on separate sets of participants, based on a 7-point Likert scale.  The full details of the selected music are described by White et al.\ \cite{white2018}.

\subsection{Autonomous Vehicle Behavior and Learning Architecture}

Our main novel contribution is focusing on whether an agent can learn a specialized policy that takes the background music into account in a way that helps it achieve its goals. To that end, during the first half (8 blocks of 12 trials, 96 trials total) the autonomous agent randomly explores while learning two models simultaneously - one that takes the background music into account (music-aware), and one that does not (music-unaware). The second phase of the experiment was partitioned in two. In one half of the exploitation phase (48 trials), the agent utilized the music-unaware model to make its decisions. In the other half, the agent utilized the music-aware model. The order of the two halves was counterbalanced across subjects. We hypothesized the additional information available to the music-aware model would lead to better performance, an expectation that was confirmed by the empirical evidence.  

\subsection{State-Action Representation}

Given that navigating through the intersection is an inherently sequential task, such that decisions at one timestep affect the agent's state in the next, the learning framework for the agent is that of reinforcement learning (RL). Sequential decision-making tasks are suitably formulated as a Markov Decision Process (MDP) \cite{sutton2018reinforcement}. Formally, an episodic MDP is a tuple
$(S, A, P, R, T)$ where $S$ is the set of states; $A$ the set of
actions, $P:S \times A \times S \rightarrow [0,1]$ is the state transition probability function where $P(s,a,s') = r$ denotes the probability of transitioning from state $s$ to state $s'$ when taking action $a$. $R:S \times A \rightarrow \mathbb{R}$ is the state-action reward function, where $R(s,a) = r$ means that taking action $a$ from state $s$ will yield reward $r$. $T$ is the set of terminal states, which end the episode.

In our experiment, the state space $S$ was composed of the autonomous car position on the $x$ axis; the participant-driven car position on the $y$ axis; the speed of the two vehicles; and the in-trial elapsed time. On top of these 5 features, 3 more features were added to the state representation to simplify learning: a binary feature for whether the autonomous vehicle reached the end of the screen; a binary feature for whether the human participant has reached the end of the screen; and a binary feature indicating whether the two vehicles crashed. Ultimately, this state representation led to an $8$-dimensional vector. The music-aware model added one more feature, a binary flag denoting whether the background music was from the ``happy'' or the ``sad'' song pool, leading to an $9$-dimensional state representation. The categorization of music as happy vs. sad was predetermined based on previous work\cite{white2018}, as we consider this task outside the scope of this study. Sentiment analysis is a well-studied field in music information retrieval and tools from this body of literature could be incorporated in future interactive systems when facing unfamiliar music\cite{liebman2020artificial}.

\begin{figure}[!htb]
\centering 
 \includegraphics[height=100pt,width=.8\columnwidth]{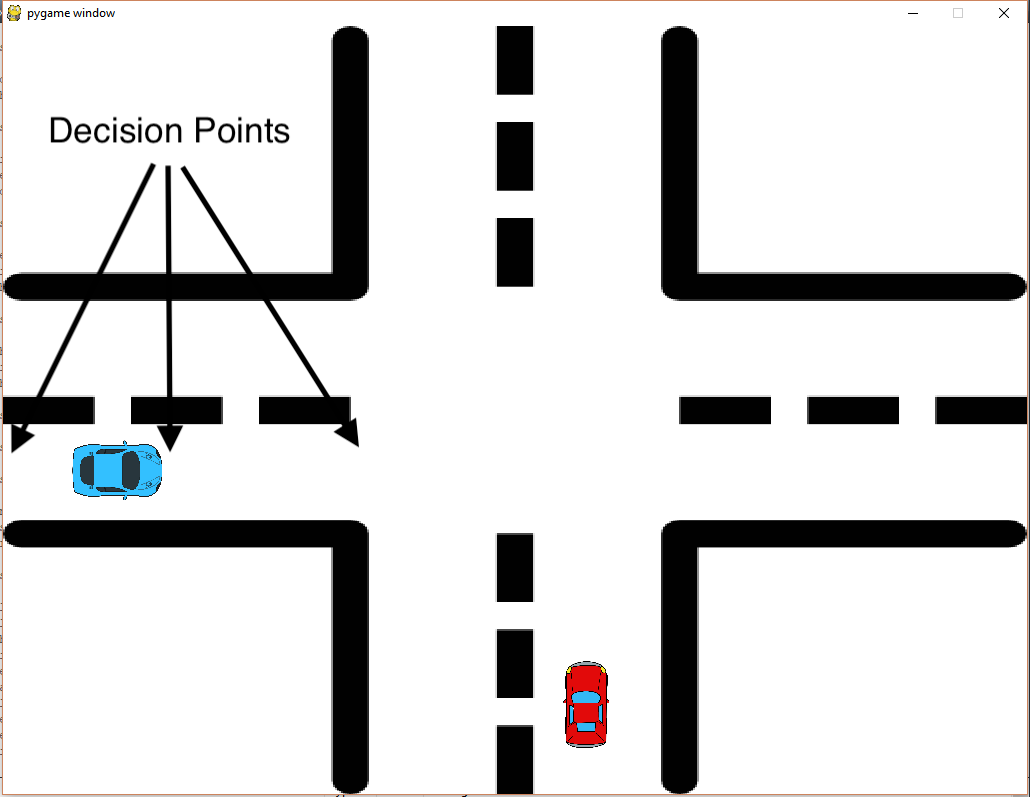}
 \caption{Decision points for the autonomous vehicle. At each point, the choices for speed are $\textbf{FAST}$, $\textbf{SLOW}$, and $\textbf{BRAKE}$.}
 \label{chap5:figdec}
\end{figure}

To simplify and standardize the decision-making process, the agent made decisions at 3 decision points: right at the start of the trial, once upon reaching the mid-point towards the intersection, and once upon reaching the intersection. If the agent stopped, it would make an additional decision after waiting for a randomized duration of 3-5 seconds. An illustration of the decision points is provided in Figure \ref{chap5:figdec}.

The action space $A$ comprised three possible actions at each decision point - $\{ \textbf{FAST}, \textbf{SLOW}, \textbf{BRAKE} \}$. An episode began with both vehicles at the starting position and ended when both vehicles reached the end (a terminal state was added at the end of each trial).

\subsubsection{Reward Function and Learning Architecture}

The goal of the agent was to reach the other end as quickly as possible without crashing. For that reason, the reward function was designed to promote expedient completion of the task, but punish the agent severely for crashing. For a completion time $t$ and a binary variable $crashed \in \{ 0,1 \}$, the reward $r$ for an episode is $r = -t - \textit{crashed} \cdot 100$ (effectively in this case maximizing reward is tantamount to minimizing penalty). 

The learning model utilized was DQN, a deep-learning variant of Q-learning, a canonical reinforcement learning technique \cite{mnih2013playing}, that has proven an effective function approximation scheme in a variety of settings. In Q-Learning, given a state-action transition $\langle s, a, r, s' \rangle$, meaning that the agent has taken action $a$ from state $s$ yielding reward $r$ and leading to new state $s'$, the expected utility of taking an action $a$ from state $s$ is learned through the update step $Q(s,a) = r + \gamma \cdot \textit{max}_{a'}(s',a')$, with $\gamma$ being the discount factor (the discount factor for the autonomous agent in this experiment was set to $\gamma = 0.9$). In DQN, the $Q$ function is approximated using a deep neural network (in our case, a feed forward network in a fashion similar to that used by Maicas et al. \cite{maicas2017deep}). In the implementation used in this experiment, a network used consisted of an input layer of sizes 8 or 9 depending on whether it was the music-aware or music-unaware model, two fully connected layers of 32 neurons each (using a ReLU activation function), and an output layer of size 3, with one output per action predicting its relative $Q$ -value. 

%
%

Since learning from limited experience in real time is impractical, the learning models utilized experience replay, a common practice in deep RL learning architectures \cite{mnih2013playing}. At the end of each trial a random sample of 20 trials (with replacement) was drawn uniformly from the aggregated trial history thus far, and the two models were trained repeatedly for 100 iterations. To further speed up learning the discounted reward was back-propagated from the end of each episode (i.e. trial) to all the $\langle s, a, s' \rangle$ elements of an episode of length $T$, i.e., for the $i$-th $\langle s, a, s' \rangle$ tuple in the episode, $r = \gamma^T-i \cdot r^T$.

In the next section we describe the results of this experiment.

\section{Results}

In this section we describe the key finding of this experiment which confirmed our main hypothesis, specifically that  utilizing a music-aware model enables better performance than the music-unaware model with respect to average task completion time, while maintaining a crash rate at least as low as that of the music-unaware model.  Completion time was calculated as the time from the beginning of the trial to when the relevant vehicle reached the end of its path. Crashes are omitted from these aggregations to avoid distortions to the measured distribution (we later show crash rates are about the same for all conditions, and we compare the aggregates unpaired).

\subsection{Impact of Learning on Average Autonomous Agent Completion Time}

\begin{figure}[!htb]
\centering 
 \includegraphics[width=\linewidth,height=180pt]{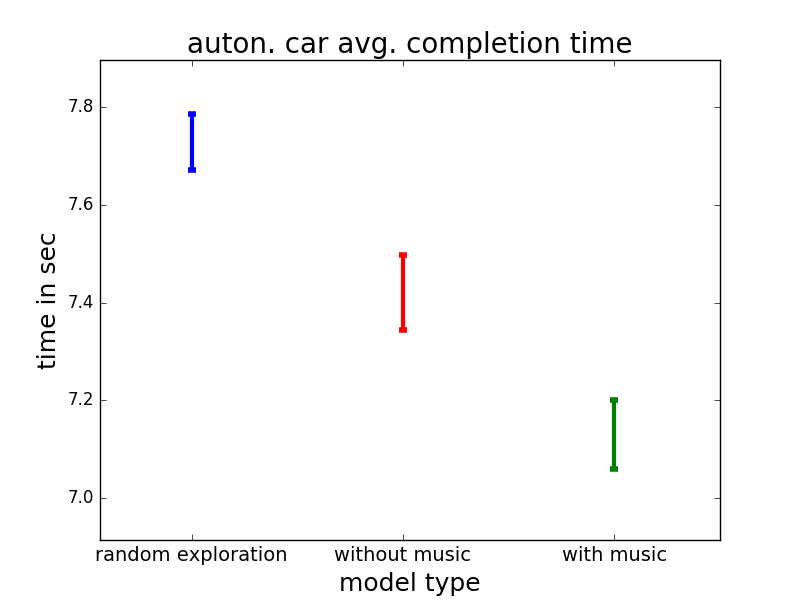}
 \caption{Average completion times for the autonomous vehicle in random exploration vs. music-unaware exploitation vs. music-aware exploitation. Error bars represent the standard error. Results are statistically significant using both an unpaired t-test and a nonparametric Mann-Whitney u-test across all condition pairs with p << 0.001.}
 \label{chap5:figtimes1}
\end{figure}

The first objective was to compare the actual completion times of the learning agent in the three stages of the experiment - exploration, exploitation without music, and exploitation with music. The results of this analysis, presented in Figure \ref{chap5:figtimes1}, first show that both learned models improved over random exploration, implying they had successfully learned some signal of the other person's intentions. Second, and more importantly, the results show that the music-aware model did significantly better than the music-unaware model, providing evidence that knowledge of the background music is indeed helpful to the learning agents. A power analysis of the core result gives an effect size value of 0.13, implying a sample size of 19-20 subjects, which means 22 participants is reasonable. However, to further confirm the robustness of these results, we repeated the analysis with a random subsample of 16 of the 22 participants 10 times and the results replicated in each run, implying they are not sensitive to the specific set of full observations. We note that while more sophisticated measures of reward exist in the human-robot collaboration literature \cite{hoffman2019evaluating}, we focus on this metric as an easily-interpretable proxy for the autonomous agent's independent utility.

\begin{figure}[!htb]
\centering 
 \includegraphics[width=.8\linewidth,height=140pt]{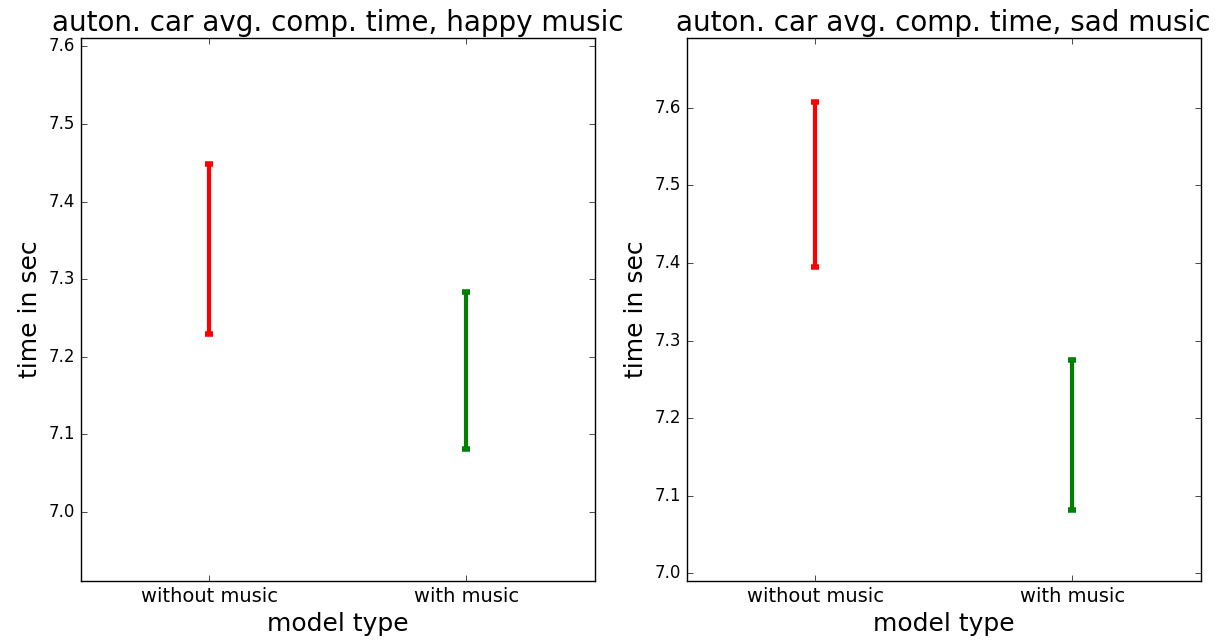}
 \caption{Average completion times for the autonomous vehicle in music-unaware exploitation vs. music-aware exploitation, split by music condition (happy music to the left, sad music to the right). Error bars represent the standard error.  This result is statistically significant using both an unpaired t-test and a nonparametric Mann-Whitney u-test with p < 0.01.}
 \label{chap5:figtimes2}
\end{figure}

 Interestingly, if we analyze the relative performance of the two models in the two background music conditions, i.e., happy vs. sad music, we can clearly see the difference in performance is much greater in the sad case, implying that the music-aware model was able to take better advantage of the participants' slower-on-average driving under this condition. This observation is presented in Figure \ref{chap5:figtimes2}.

\begin{figure}[!htb]
\centering 
 \includegraphics[width=.8\linewidth,height=140pt]{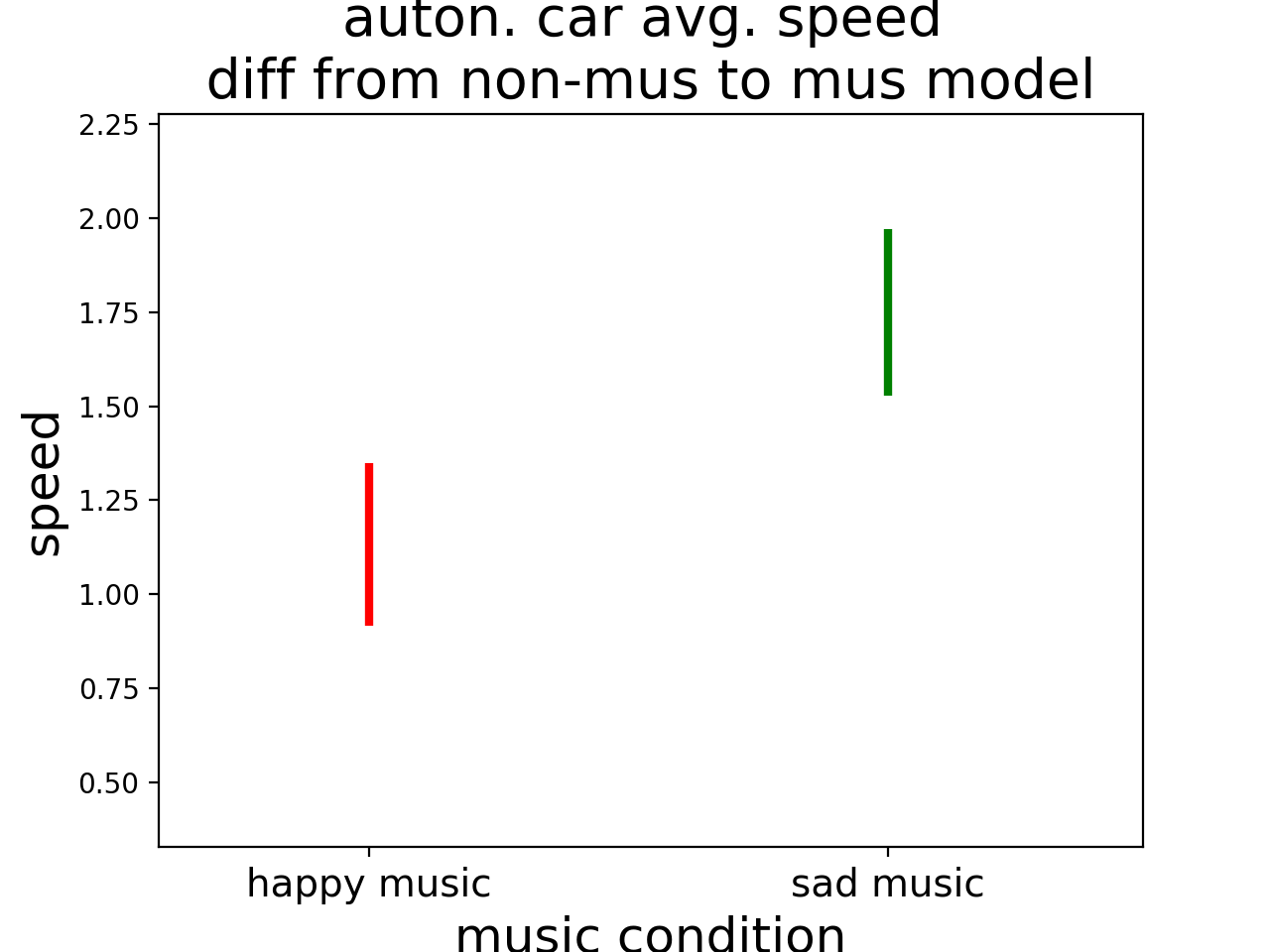}
 \caption{Average difference in autonomous driving speed between the music-unaware vs. music-aware models. The difference is greater with the background condition of sad music.  This result is statistically significant using both an unpaired t-test and a nonparametric Mann-Whitney u-test with p << 0.001}
 \label{chap5:speeddiff}
\end{figure}

Indeed, if we consider the differential effect of the background music on driving speed in the two learned models, we can see that while the music model leads to the autonomous vehicle driving faster on average, the effect is statistically significantly greater in the sad music condition relative to the happy music condition. This result is highlighted in Figure \ref{chap5:speeddiff}.

%
%

\subsection{Impact of Background Music on Learned Policies}
\label{impact_policies}
Having established that there is indeed a demonstrable difference in performance between the music aware and the music agnostic learned policies, it is of interest to dig a bit deeper into what differentiates these two policies with respect to autonomous car behavior. For that purpose, we consider the distributions of actions  induced by each policy.  To this end, we consider the empirical likelihood of taking each of the three actions ($\textbf{SLOW}$, $\textbf{FAST}$ and $\textbf{BRAKE}$ under the two conditions (happy vs. sad background music), for each of the two models. 

\begin{figure}[!htb]
\centering 
 \includegraphics[width=\linewidth, height=130pt]{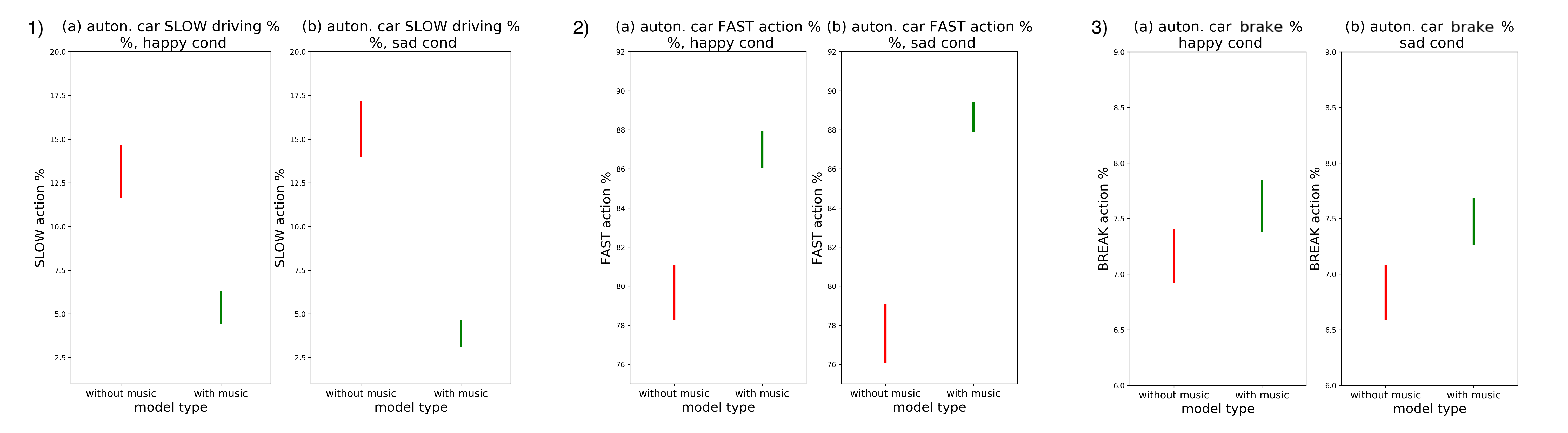}
 \caption{Difference in frequency of taking each action in the happy vs. sad music condition. 1) $\textbf{SLOW}$; 2) $\textbf{FAST}$; 3) $\textbf{BRAKE}$.  For each figure (a) represents the happy music condition, and figure (b) represents the sad music condition.}
 \label{slowp}
\end{figure}

Interestingly, the music-aware model uses substantially fewer $\textbf{SLOW}$ actions and substantially more $\textbf{FAST}$ actions than the music-aware model, as observed in Figure \ref{slowp}. Corroborating the findings presented in Figure \ref{chap5:speeddiff}, the difference between action frequencies in the music-aware vs.\ music-unaware models is substantially greater in the sad music condition. However, somewhat counter-intuitively, the likelihood of braking actually goes up in the music-aware vs. the music-unaware model. This observation can be explained as a form of balance - since the model drives the car faster overall, it has to brake somewhat more often to avoid crashing. Indeed, in the next section we study the impact of using music as part of the input on the crash rates directly.

\begin{figure}[!htb]
\centering 
 \includegraphics[width=\linewidth, height=250pt]{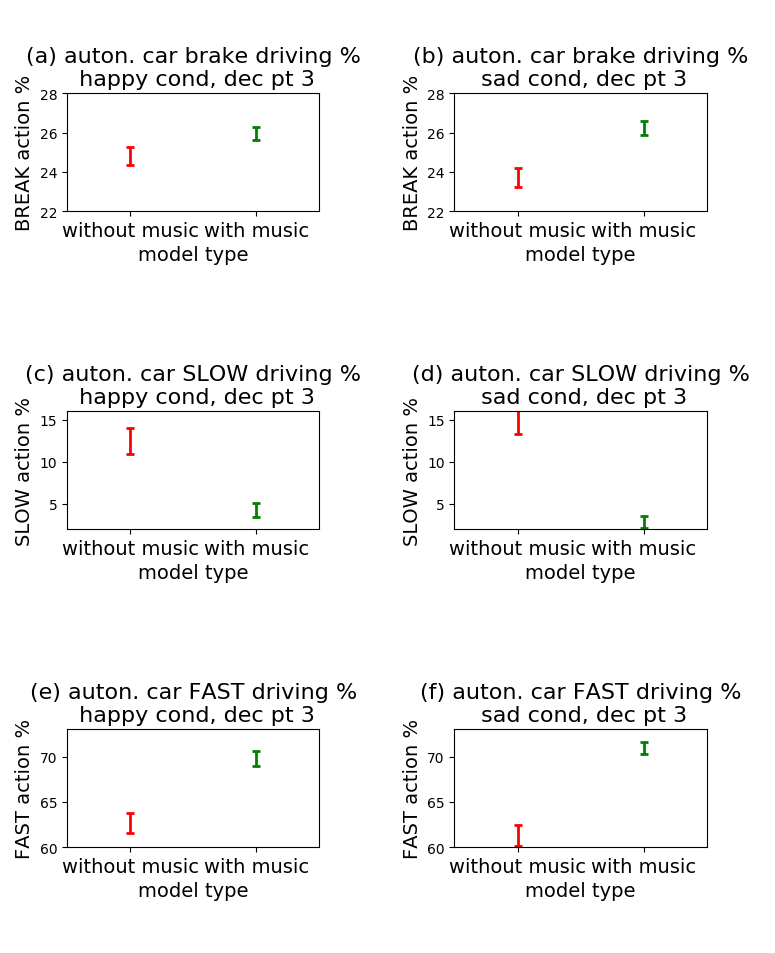}
 \caption{Differences in frequency for taking each action with the music-unaware vs. music aware models in the third (intersection) decision point with both music conditions. Figures a,b cover the $\textbf{BRAKE}$ action; Figures c,d cover the $\textbf{SLOW}$ action; Figures e,f cover the $\textbf{FAST}$ action. Figures a,c,e cover the happy music condition; Figures b,d,f cover the sad music condition.}
 \label{decpoint_hs}
\end{figure}

This difference nicely highlights how the policy differences are decomposed by state context. It is of note that neither model took the $\textbf{BRAKE}$ action in the first and second decision points after learning, as expected. However, supporting the observation presented in Figure \ref{slowp}, we can see that the music-aware model brakes significantly more often than the music-unaware model, but opts for the $\textbf{SLOW}$ driving action drastically less often. Interestingly, additional breakdown by type of background music (happy vs.\ sad) does not reveal statistically significant differences between the models with respect to the combination of action \emph{and} music condition for the first two decision points, but there were statistically significant differences in all three actions at the third (intersection) decision point, as illustrated in Figure \ref{decpoint_hs}.

\subsection{Impact on the Crash Rate}

\begin{figure}[!htb]
\centering 
 \includegraphics[width=\columnwidth,height=150pt]{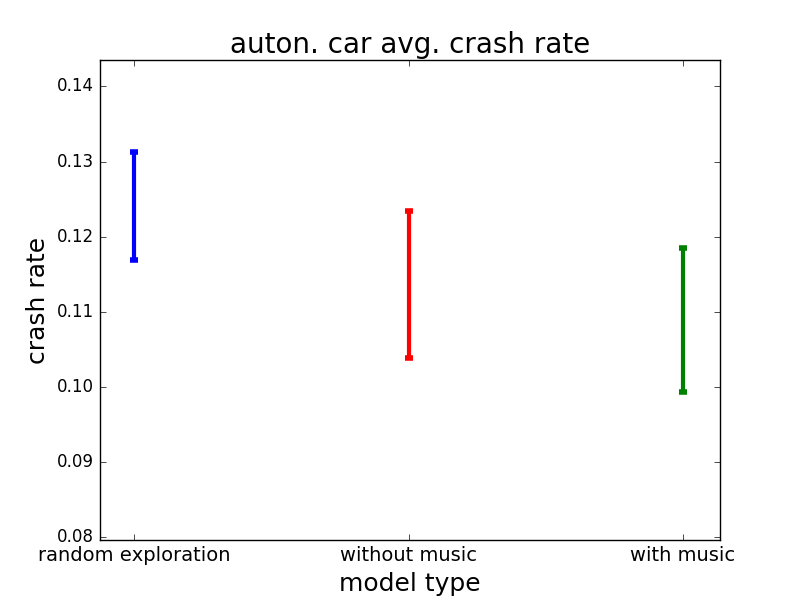}
 \caption{Average crash rates for the autonomous vehicle in exploration vs. music-unaware expoitation vs. music-aware exploitation, split by music condition (happy music to the left, sad music to the right). Error bars represent the standard error.}
 \label{chap5:crashrate}
\end{figure}

One caveat when claiming one model did better than another with respect to average completion time is that if one model behaves more recklessly, it would make it substantially more crash-prone. The goal was to have the agents learn how to reach the other side quickly, but also  \emph{safely}, meaning it had to take the human participant into account. This concern is particularly salient given the differential impact using the music background condition had on the action distribution of the learned models, as indicated in Section \ref{impact_policies}. For this purpose we also examined the average crash rate in the three phases of the experiment, i.e. exploration, exploitation without music, and exploitation with music. Though the results are not statistically significant, they are sufficient to assuage the suspicion that the music-aware model did better than the music-unaware model with respect to average completion time merely by driving recklessly (if anything, the results seem to suggest an opposite trend). The comparison of crash rates in the three phases is presented in Figure \ref{chap5:crashrate}.

\section{Summary \& Discussion}

In this paper, we show evidence that music has an impact on human decision-making in the context of human-robot interaction, and that this impact can be modeled -- with a more advanced level of autonomous agency, towards engendering human-robot and robot-robot cooperation in musical environments. We did so in the context of an intersection-crossing task presented in the first experiment. A learning agent was introduced in lieu of the simpler preprogrammed agent utilized in the first experiment. This agent spent the first half of the experiment gathering data and learning two separate behavior models - one which included the type of music in the world state representation, and one that did not. In the latter half of the experiment, the agent utilized both models to decide on its actions. 

Results of this study indicated that indeed explicitly modeling a human's background music when interacting with an agent as part of the world representation helped the agent improve its performance in terms of completion time, without increasing the risk of crashing. Though these results were meant primarily as a proof of concept, the studied setting is not entirely unrealistic. Consider a future in which people are encouraged to sign up for services which share their driving information with autonomous vehicles towards better system performance (similar to driving safety apps mandated by many insurance companies). Of course, given privacy concerns, we do not expect that everyone will want to share the music they're listening to in real time.  On the other hand, some people may indeed opt in, perhaps due to the potential benefits demonstrated in this paper, or simply in accordance with their usual social media habits. Regardless of regime, in such a setting, background music could be part of the shared information.  However, the main findings of this paper also have potential implications in many other arenas of human-agent interaction, as listed in Section \ref{exp_setup}.

This result establishes that human-robot interaction in the context of music is possible and potentially beneficial, and opens the door for many potential follow-up experiments. Such follow-up work includes, but is not limited to, studying how well the findings of this paper generalize in other experimental settings; conducting similar experiments with a wider range of music; and having the autonomous agent learn a more refined representation of the background music and its predicted effects on people's behavior. This paper also raises important ethical questions involving the legitimacy of using background information in human-robot interaction and how it exposes people to subtle forms of manipulation. While this discussion goes beyond the scope of this paper, it is an important discussion for the community to engage in if we are to continue pursuing this line of research.


\begin{thebibliography}{10}
\providecommand{\url}[1]{#1}
\csname url@samestyle\endcsname
\providecommand{\newblock}{\relax}
\providecommand{\bibinfo}[2]{#2}
\providecommand{\BIBentrySTDinterwordspacing}{\spaceskip=0pt\relax}
\providecommand{\BIBentryALTinterwordstretchfactor}{4}
\providecommand{\BIBentryALTinterwordspacing}{\spaceskip=\fontdimen2\font plus
\BIBentryALTinterwordstretchfactor\fontdimen3\font minus
  \fontdimen4\font\relax}
\providecommand{\BIBforeignlanguage}[2]{{%
\expandafter\ifx\csname l@#1\endcsname\relax
\typeout{** WARNING: IEEEtran.bst: No hyphenation pattern has been}%
\typeout{** loaded for the language `#1'. Using the pattern for}%
\typeout{** the default language instead.}%
\else
\language=\csname l@#1\endcsname
\fi
#2}}
\providecommand{\BIBdecl}{\relax}
\BIBdecl

\bibitem{elliott2002neural}
R.~Elliott, J.~S. Rubinsztein, B.~J. Sahakian, and R.~J. Dolan, ``The neural
  basis of mood-congruent processing biases in depression,'' \emph{Archives of
  general psychiatry}, vol.~59, no.~7, pp. 597--604, 2002.

\bibitem{olafson2001effects}
K.~M. Olafson and F.~R. Ferraro, ``Effects of emotional state on lexical
  decision performance,'' \emph{Brain and Cognition}, vol.~45, no.~1, pp.
  15--20, 2001.

\bibitem{decbias}
E.~Liebman, P.~Stone, and C.~N. White, ``How music alters decision making -
  impact of music stimuli on emotional classification,'' in \emph{Proceedings
  of the 16th International Society for Music Information Retrieval Conference,
  {ISMIR} 2015, M{\'{a}}laga, Spain, October 26-30, 2015}, 2015, pp. 793--799.

\bibitem{liebman2016impact}
------, ``Impact of music on decision making in quantitative tasks.'' in
  \emph{ISMIR}, 2016, pp. 661--667.

\bibitem{ISMIR2018}
E.~Liebman, C.~N. White, and P.~Stone, ``On the impact of music on decision
  making in cooperative tasks,'' in \emph{ISMIR}, 2018.

\bibitem{jeong2011congruence}
J.-W. Jeong, V.~A. Diwadkar, C.~D. Chugani, P.~Sinsoongsud, O.~Muzik, M.~E.
  Behen, H.~T. Chugani, and D.~C. Chugani, ``Congruence of happy and sad
  emotion in music and faces modifies cortical audiovisual activation,''
  \emph{NeuroImage}, vol.~54, no.~4, pp. 2973--2982, 2011.

\bibitem{de2002effectiveness}
S.~K. de~l’Etoile, ``The effectiveness of music therapy in group
  psychotherapy for adults with mental illness,'' \emph{The Arts in
  Psychotherapy}, vol.~29, no.~2, pp. 69--78, 2002.

\bibitem{kuhbandner2013joint}
C.~Kuhbandner and R.~Pekrun, ``Joint effects of emotion and color on memory.''
  \emph{Emotion}, vol.~13, no.~3, p. 375, 2013.

\bibitem{blood2001intensely}
A.~J. Blood and R.~J. Zatorre, ``Intensely pleasurable responses to music
  correlate with activity in brain regions implicated in reward and emotion,''
  \emph{Proceedings of the National Academy of Sciences}, vol.~98, no.~20, pp.
  11\,818--11\,823, 2001.

\bibitem{paquette2013musical}
S.~Paquette, I.~Peretz, and P.~Belin, ``The “musical emotional bursts”: a
  validated set of musical affect bursts to investigate auditory affective
  processing,'' \emph{Frontiers in psychology}, vol.~4, 2013.

\bibitem{zumbansen2014combination}
A.~Zumbansen, I.~Peretz, and S.~H{\'e}bert, ``The combination of rhythm and
  pitch can account for the beneficial effect of melodic intonation therapy on
  connected speech improvements in broca’s aphasia,'' \emph{Frontiers in
  human neuroscience}, vol.~8, 2014.

\bibitem{chen2008music}
J.~Chen, J.~Yuan, H.~Huang, C.~Chen, and H.~Li, ``Music-induced mood modulates
  the strength of emotional negativity bias: An erp study,'' \emph{Neuroscience
  Letters}, vol. 445, no.~2, pp. 135--139, 2008.

\bibitem{spenwyn2010role}
J.~Spenwyn, D.~J. Barrett, and M.~D. Griffiths, ``The role of light and music
  in gambling behaviour: An empirical pilot study,'' \emph{International
  Journal of Mental Health and Addiction}, vol.~8, no.~1, pp. 107--118, 2010.

\bibitem{griffiths2005psychology}
M.~Griffiths and J.~Parke, ``The psychology of music in gambling environments:
  An observational research note,'' \emph{Journal of Gambling Issues}, 2005.

\bibitem{dixon2007empirical}
L.~Dixon, R.~Trigg, and M.~Griffiths, ``An empirical investigation of music and
  gambling behaviour,'' \emph{International Gambling Studies}, vol.~7, no.~3,
  pp. 315--326, 2007.

\bibitem{noseworthy2009comparison}
T.~J. Noseworthy and K.~Finlay, ``A comparison of ambient casino sound and
  music: Effects on dissociation and on perceptions of elapsed time while
  playing slot machines,'' \emph{Journal of Gambling Studies}, vol.~25, no.~3,
  pp. 331--342, 2009.

\bibitem{nayak2000effect}
S.~Nayak, B.~L. Wheeler, S.~C. Shiflett, and S.~Agostinelli, ``Effect of music
  therapy on mood and social interaction among individuals with acute traumatic
  brain injury and stroke.'' \emph{Rehabilitation Psychology}, vol.~45, no.~3,
  p. 274, 2000.

\bibitem{greitemeyer2011exposure}
T.~Greitemeyer, ``Exposure to music with prosocial lyrics reduces aggression:
  First evidence and test of the underlying mechanism,'' \emph{Journal of
  Experimental Social Psychology}, vol.~47, no.~1, pp. 28--36, 2011.

\bibitem{baron1990environmentally}
R.~A. Baron, ``Environmentally induced positive affect: Its impact on
  self-efficacy, task performance, negotiation, and conflict,'' \emph{Journal
  of Applied Social Psychology}, vol.~20, no.~5, pp. 368--384, 1990.

\bibitem{white2018}
\BIBentryALTinterwordspacing
C.~N. White, E.~Liebman, and P.~Stone, ``Decision mechanisms underlying
  mood-congruent emotional classification,'' \emph{Cognition and Emotion},
  vol.~32, no.~2, pp. 249--258, 2018, pMID: 28271732. [Online]. Available:
  \url{https://doi.org/10.1080/02699931.2017.1296820}
\BIBentrySTDinterwordspacing

\bibitem{sutton2018reinforcement}
R.~S. Sutton and A.~G. Barto, \emph{Reinforcement learning: An
  introduction}.\hskip 1em plus 0.5em minus 0.4em\relax MIT press, 2018.

\bibitem{mnih2013playing}
V.~Mnih, K.~Kavukcuoglu, D.~Silver, A.~Graves, I.~Antonoglou, D.~Wierstra, and
  M.~Riedmiller, ``Playing atari with deep reinforcement learning,''
  \emph{arXiv preprint arXiv:1312.5602}, 2013.

\bibitem{maicas2017deep}
G.~Maicas, G.~Carneiro, A.~P. Bradley, J.~C. Nascimento, and I.~Reid, ``Deep
  reinforcement learning for active breast lesion detection from dce-mri,'' in
  \emph{International Conference on Medical Image Computing and
  Computer-Assisted Intervention}.\hskip 1em plus 0.5em minus 0.4em\relax
  Springer, 2017, pp. 665--673.

\bibitem{hoffman2019evaluating}
Hoffman, G. Evaluating fluency in human–robot collaboration. {\em IEEE Transactions On Human-Machine Systems}. \textbf{49}, 209-218 (2019)

\bibitem{liebman2020artificial}
Liebman, E. \& Stone, P. Artificial musical intelligence: A survey. {\em ArXiv Preprint ArXiv:2006.10553}. (2020)

\bibitem{thomaz2016computational}Thomaz, A., Hoffman, G. \& Cakmak, M. Computational human-robot interaction. {\em Foundations And Trends In Robotics}. \textbf{4}, 105-223 (2016)

\bibitem{liebman2020sequential}Liebman, E. Sequential decision-making in musical intelligence. (Springer,2020)


\end{thebibliography}
\end{document}